\documentclass[12pt,draftcls,onecolumn]{IEEEtran}
\IEEEoverridecommandlockouts 
\usepackage{amssymb,amsmath,color,psfrag}
\usepackage{graphicx}
\usepackage{algorithm,algorithmic,xspace}

\renewcommand{\natural}{{\mathbb{N}}}

\newcommand{\real}{{\mathbb{R}}}
\newcommand{\subscr}[2]{{#1}_{\textup{#2}}}

\newcommand{\map}[3]{#1: #2 \rightarrow #3}
\newcommand{\setdef}[2]{\{#1 \; | \; #2\}}

\newcommand{\DD}{\mathcal{D}}
\newcommand{\PP}{\mathcal{P}}
\newcommand{\umax}{\subscr{r}{ctr}}
\newcommand{\vmax}{\subscr{r}{vel}}
\newcommand{\pointset}{P}

\newcommand{\setpointsets}[1]{\PP_{#1}}
\newcommand{\domain}{\ensuremath{\mathcal{Q}}}
\newcommand{\ETSP}[1]{\operatorname{ETSP}(#1)}
\newcommand{\SOTSP}[1]{\operatorname{DITSP}(#1)}

\newcommand{\diam}{\operatorname{diam}}
\newcommand{\norm}[1]{\|#1\|}
\newcommand{\maximize}{\text{maximize}}
\newcommand{\subj}{\text{subj. to}}
\newcommand{\E}{\operatorname{E}}

\newcommand{\Length}{\operatorname{Length}}
\newcommand{\Area}{\operatorname{Area}}
\newcommand{\Volume}{\operatorname{Volume}}

\newcommand{\DTSP}[2]{\operatorname{DTSP}_{#2}(#1)}
\newcommand{\AltAlgo}{\textsc{Alternating Algorithm}\xspace}
\newcommand{\BeadTilingAlgo}{\textsc{Bead-Tiling Algorithm}\xspace}
\newcommand{\BTA}{\textsc{Bead Tiling Algorithm}\xspace}
\newcommand{\CFA}{\textsc{Cylinder Covering Algorithm}\xspace}
\newcommand{\RecBTA}{\textsc{RecBTA}\xspace}
\newcommand{\RecCCA}{\textsc{RecCCA}\xspace}
\newcommand{\shortBTA}{\textsc{BTA}\xspace}
\newcommand{\shortCCA}{\textsc{CCA}\xspace}
\newcommand{\RecBeadTilingAlgo}{\textsc{Recursive Bead-Tiling Algorithm}\xspace}
\newcommand{\RecCylFillingAlgo}{\textsc{Recursive Cylinder-Covering Algorithm}\xspace}
\newcommand{\LenRBTA}[2]{\operatorname{L_{\mathrm{RBTA}, #2}}(#1)}
\newcommand{\LenRCFA}[2]{\operatorname{L_{\mathrm{RCFA}, #2}}(#1)}

\newcommand{\tile}{\ensuremath{\mathcal{B}_{\rho}}}
\newcommand{\cyl}{\ensuremath{\mathcal{C}_{\rho}}}

\newcommand{\DoubIntTSP}{\text{DITSP}}
\newcommand{\width}{W}
\newcommand{\height}{H}
\newcommand{\depth}{D}

\newtheorem{theorem}{Theorem}[section]

\newtheorem{lemma}[theorem]{Lemma}
\newtheorem{remark}[theorem]{Remark}

\def\QEDclosed{\mbox{\rule[0pt]{1.3ex}{1.3ex}}} 

\def\QED{\QEDclosed} 

\newcommand\oprocendsymbol{\hbox{$\square$}}
\newcommand\oprocend{\relax\ifmmode\else\unskip\hfill\fi\oprocendsymbol}

\renewcommand{\theenumi}{(\roman{enumi})}

\begin{document}

\title{\LARGE \bf Traveling Salesperson Problems for a double integrator}

\author{Ketan Savla \and Francesco Bullo\thanks{Ketan Savla and Francesco
    Bullo are with the Center for Control, Dynamical Systems and Computation,
    University of California at Santa Barbara,
    \texttt{ketansavla@umail.ucsb.edu, bullo@engineering.ucsb.edu}} \and
 Emilio Frazzoli\thanks{Emilio Frazzoli is with the Aeronautics and Astronautics Department, Massachusetts Institute of Technology,  \texttt{frazzoli@mit.edu}}}
  
\maketitle

\begin{abstract}
  In this paper we propose some novel path planning strategies for a double
  integrator with bounded velocity and bounded control inputs. First, we
  study the following version of the Traveling Salesperson Problem (TSP):
  given a set of points in $\real^d$, find the fastest tour over the point
  set for a double integrator. We first give asymptotic bounds on the time
  taken to complete such a tour in the worst-case. Then, we study a
  stochastic version of the TSP for double integrator where the points are
  randomly sampled from a uniform distribution in a compact environment in
  $\real^2$ and $\real^3$. We propose novel algorithms that perform within
  a constant factor of the optimal strategy with high probability. Lastly,
  we study a dynamic TSP: given a stochastic process that generates
  targets, is there a policy which guarantees that the number of unvisited
  targets does not diverge over time? If such stable policies exist, what
  is the minimum wait for a target?  We propose novel stabilizing
  receding-horizon algorithms whose performances are within a constant
  factor from the optimum with high probability, in $\real^2$ as well as
  $\real^3$. We also argue that these algorithms give identical
  performances for a particular nonholonomic vehicle, Dubins vehicle.
\end{abstract}  

\section{Introduction}
The Traveling Salesperson Problem (TSP) with its variations is one of the
most widely known combinatorial optimization problems. While extensively
studied in the literature, these problems continue to attract great
interest from a wide range of fields, including Operations Research,
Mathematics and Computer Science. The Euclidean TSP (ETSP)
\cite{JB-JH-JH:59,JMS:90} is formulated as follows: given a finite point
set $\pointset$ in $\real^d$ for $d \in \natural$, find the minimum-length
closed path through all points in $\pointset$.  It is quite natural to
formulate this problem in the context of other dynamical vehicles.  The
focus of this paper is the analysis of the TSP for a vehicle with double
integrator dynamics or simply a double integrator; we shall refer to it as
$\DoubIntTSP$. Specifically, $\DoubIntTSP$ will involve finding the
\emph{fastest} tour for a double integrator through a set of points.

Exact algorithms, heuristics and polynomial-time constant factor
approximation algorithms are available for the Euclidean TSP,
see~\cite{DA-RB-VC-WC:98,SA:97,SL-BWK:73}. However, unlike most other
variations of the TSP, it is believed that the $\DoubIntTSP$ cannot be
formulated as a problem on a finite-dimensional graph, thus preventing the
use of well-established tools in combinatorial optimization. On the other
hand, it is reasonable to expect that exploiting the geometric structure of
feasible paths for a double integrator, one can gain insight into the
nature of the solution, and possibly provide polynomial-time approximation
algorithms.

The motivation to study the $\DoubIntTSP$ arises in robotics and
uninhabited aerial vehicles (UAVs) applications. In particular, we envision
applying our algorithm to the setting of an UAV monitoring a collection of
spatially distributed points of interest. Additionally, from a purely
scientific viewpoint, it is of general interest to bring together the work
on dynamical vehicles and that on TSP. UAV applications also motivate us to
study the Dynamic Traveling Repairperson Problem (DTRP), in which the
aerial vehicle is required to visit a dynamically generated set of targets.
This problem was introduced by Bertsimas and van Ryzin in
\cite{DJS-GJvR:91} and then decentralized policies achieving the same
performances were proposed in \cite{EF-FB:03r}. Variants of these problems
have attracted much attention recently
\cite{EF-FB:03r,ZT-UO:05,RWB-TWM-MAG-EPA:02,SD:05,SR-RS-SD:05}. However, as
with the TSP, the study of DTRP in conjunction with vehicle dynamics has
eluded attention from the research community.

The contributions of this paper are threefold.  First, we analyze the minimum time to traverse
$\DoubIntTSP$ in $\real^d$ for $d \in \natural$. We show that the minimum time to traverse
$\DoubIntTSP$ belongs to $O(n^{1-\frac{1}{2d}})$ and in the worst case,
it also belongs\footnote{For
  $\map{f,g}{\natural}{\real}$, we say that $f \in O(g)$ (respectively, $f
  \in \Omega(g)$) if there exist $N_0 \in \natural$ and $k \in \real_+$
  such that $|f(N)| \le k |g(N)|$ for all $N \ge N_0$ (respectively,
  $|f(N)| \ge k |g(N)|$ for all $N \ge N_0$).  If $f \in O(g)$ and $f \in
  \Omega(g)$, then we use the notation $f \in \Theta (g)$.} to $\Omega(n^{1-\frac{1}{d}})$.  Second, we
study the \emph{stochastic} $\DoubIntTSP$, i.e., the problem of finding the
fastest tour through a set of target points that are uniformly randomly
generated. We show that the minimum time to traverse the tour for the stochastic $\DoubIntTSP$ belongs to
$\Omega(n^{2/3})$ in $\real^2$ and $\Omega(n^4/5)$ in $\real^3$.  Drawing
inspiration from our earlier work~\cite{KS-EF-FB:06h-tmp}, we propose two
novel algorithms for the stochastic $\DoubIntTSP$: the \RecBeadTilingAlgo
for $\real^2$ and the \RecCylFillingAlgo for $\real^3$.  We prove that
these algorithms provide a constant-factor approximation to the optimal
$\DoubIntTSP$ solution with high probability.  Third, we propose two
algorithms for the DTRP in the heavy load case based on the
fixed-resolution versions of the corresponding algorithms for stochastic
$\DoubIntTSP$. We show that the performance guarantees for the stochastic
$\DoubIntTSP$ translate into stability guarantees for the average
performance of the DTRP problem for a double integrator.  Specifically, the
performances of the algorithms for the DTRP are within a constant factor of
the optimal policies.  We contend that the successful application to the
DTRP problem does indeed demonstrate the significance of the $\DoubIntTSP$
problem from a control viewpoint. As a final minor contribution, we also
show that the results obtained for stochastic $\DoubIntTSP$ carry over to
the stochastic TSP for the Dubins vehicle, i.e., for a nonholonomic vehicle
moving along paths with bounded curvature, without reversing direction.

This work completes the generalization of the known combinatorial results
on the ETSP and DTRP (applicable to systems with single integrator
dynamics) to double integrators and Dubins vehicle models.  It is
interesting to compare our results with the setting where the vehicle is
modeled by a single integrator; this setting corresponds to the so-called
Euclidean case in combinatorial optimization. The results are summarized as
follows: \smallskip

\begin{center}
\noindent\resizebox{.63\linewidth}{!}{
\begin{tabular}{|c|c|c|c|}
\hline
&Single& Double& Dubins\\
&integrator & integrator & vehicle \\
\hline \hline
Min. time for & $\Theta(n^{1-\frac{1}{d}})$ \cite{JMS:90} & $\Omega(n^{1-\frac{1}{d}})$, & $\Theta(n)$ \cite{KS-EF-FB:04l}\\
TSP tour & & $O(n^{1-\frac{1}{2d}})$& ($d=2,3$)\\
(worst-case) & & &\\
\hline
Exp. min. time & $\Theta(n^{1-\frac{1}{d}})$ \cite{JMS:90} & $\Theta(n^{1-\frac{1}{2d-1}})$ & $\Theta(n^{1-\frac{1}{2d-1}})$\\
for TSP tour & & w.h.p.& w.h.p. \\
(stochastic) & &($d=2,3$)&($d=2,3$)\\
\hline
System time & $\Theta(\lambda^{d-1})$ \cite{DJS-GJvR:91}& $\Theta(\lambda^{2(d-1)})$ & $\Theta(\lambda^{2(d-1)})$\\
for DTRP & ($d=1$)&($d=2,3$)&($d=2,3$) \\
\hline
\end{tabular}}
\end{center}


\smallskip Remarkably, the differences between these various bounds for the
TSP play a crucial role when studying the DTRP problem; e.g., stable
policies exist only when the minimum time for traversing the TSP tour grows
strictly sub-linearly with $n$.  For the DTRP problem we propose novel
policies and show their stability for a uniform target-generation process
with intensity $\lambda$.  It is clear from the table that motion
constraints make the system much more sensitive to increases in the target
generation rate $\lambda$.

\section{Setup and worst-case DITSP}
For $d \in \natural$, consider a vehicle with double integrator dynamics:
\begin{equation}
  \ddot{p}(t)=u(t),\quad \| u(t)\| \leq \umax, \quad \| \dot{p}(t)\| \leq \vmax, 
  \label{eq:second-order-model}
\end{equation}
where $p \in \real^d$ and $u \in \real^d$ are the position and control
input of the vehicle, $\vmax \in \real_+$ and $\umax \in \real_+$ are the
bounds on the attainable speed and control inputs.  Let
$\domain\subset\real^d$ be a unit hypercube.  Let
$\pointset=\{q_1,\ldots,q_n\}$ be a set of $n$ points in $\domain$ and
$\setpointsets{n}$ be the collection of all point sets $\pointset \subset
\domain$ with cardinality $n$.  Let $\ETSP{\pointset}$ denote the cost of
the Euclidean TSP over $\pointset$ and let $\SOTSP{\pointset}$ denote the
cost of the TSP for double integrator over $\pointset$, i.e., the time
taken to traverse the fastest closed path for a double integrator through
all points in $\pointset$. We assume $\vmax$ and $\umax$ to be constant and
we study the dependence of $\map{$\DoubIntTSP$}{\setpointsets{n}}{\real_+}$
on $n$.  Without loss of generality, we assume the vehicle starts
traversing the TSP tour at $t=0$ with initial position~$q_1$.
\begin{lemma}\textit{(Worst-case Lower Bound on the TSP for Double Integrator)}
  \label{lem:lower-bound-deterministic}
  For $\vmax,\umax\in\real_+$ and $d \in \natural$, there exists a point
  set $\pointset \in \setpointsets{n}$ in $\domain \subset \real^d$ such
  that $\SOTSP{\pointset}$ belongs to $\Omega(n^{1-\frac{1}{d}})$.
\end{lemma}
\begin{proof}
  We consider the class of point sets that give rise to the worst case
  scenario for the ETSP; we refer the reader to~\cite{JMS:90}.  It suffice
  to note that, for such a point set of cardinality $n$ in $\real^d$, the
  minimum distance between any two points belongs to
  $\Omega(n^{-\frac{1}{d}})$. The minimum time required for a double
  integrator with initial speed $\tilde{v}$ to go from one point to another
  at a distance $\tilde{\delta}$ is lower bounded by
  $\sqrt{(\tilde{v}/\umax)^2+2 (\tilde{\delta}/\umax)}-\tilde{v}/\umax$.
  However, $\tilde{v} \leq \vmax$ and for the point set under
  consideration, $\tilde{\delta}$ belongs to $\Omega(n^{-\frac{1}{d}})$.
  This implies that the minimum time required for a double integrator to
  travel between two points of the given point set belongs to
  $\Omega(n^{-\frac{1}{d}})$.  Hence, the minimum time required for the
  vehicle to complete the tour over this point set belongs to $n
  \Omega(n^{-\frac{1}{d}})$, i.e., $\Omega(n^{1-\frac{1}{d}})$.
\end{proof}
 
We now propose a simple strategy for the $\DoubIntTSP$ and analyze its
performance. The STOP-GO-STOP strategy can be described as follows: The
vehicle visits the points in the same order as in the optimal ETSP tour
over the same set of points.  Between any pair of points, the vehicle
starts at the initial point at rest, follows the shortest-time path to
reach the final point with zero velocity.

\begin{theorem}\textit{(Upper Bound on the TSP for Double Integrator)}
  \label{theorem:upper-bound-deterministic}
  For any point set $\pointset \in \setpointsets{n}$ in $\domain \subset
  \real^d$ and $\umax > 0$, $\vmax > 0$ and $d \in \natural$,
  $\SOTSP{\pointset}$ belongs to $O(n^{1-\frac{1}{2d}})$.
\end{theorem}
\begin{proof}
  Without any loss of generality, let
  $(q_1,\ldots,q_n,q_1)$ be the optimal order of points for the Euclidean
  TSP over $\pointset$.  For $1 \leq i \leq n-1$, let
  $\delta_i=\norm{q_i-q_{i+1}}$ and $\delta_n=\norm{q_n-q_1}$. If
  $\delta_i$ is the distance between a set of points, then the time $t_i$
  required to traverse that distance by a double integrator following the
  STOP-GO-STOP strategy is given by:
  \begin{equation*}
    t_i=  \begin{cases}
      2\sqrt{\frac{\delta_i}{\umax}}, & \quad\textrm{if $\delta_i \leq
      \frac{\vmax^2}{\umax}$},\\      
    \frac{\vmax}{\umax} + \frac{\delta_i}{\vmax}, & \quad
      \textrm{otherwise}. 
    \end{cases}
  \end{equation*}
  Let $\mathcal{I}=\setdef{1 \leq i \leq n}{\delta_i \leq \vmax^2/\umax}$
  and $\mathcal{I}^c=\{1,\ldots,n\} \setminus \mathcal{I}$. Also, let
  $n_{\mathcal{I}}$ be the cardinality of the set $\mathcal{I}$ and let
  $n_{\mathcal{I}^c}=n-n_{\mathcal{I}}$.  Therefore, an upper bound on the
  minimum time taken to complete the tour as obtained from this strategy is
  \begin{equation}\label{eq:first-expr-for-SOTSP}
    \begin{split}
      \SOTSP{\pointset} & \leq \sum_{i=1}^n t_i = \sum_{i \in \mathcal{I}} t_i + \sum_{i \in \mathcal{I}^c} t_i  
      =\frac{2}{\sqrt{\umax}} \sum_{i \in \mathcal{I}} \sqrt{\delta_i} +
      n_{\mathcal{I}^c} \frac{\vmax}{\umax} + \frac{1}{\vmax} \sum_{i \in
      \mathcal{I}^c} \delta_i  \\ 
    &\leq \frac{2}{\sqrt{\umax}} \sum_{i \in \mathcal{I}} \sqrt{\delta_i} +
      n_{\mathcal{I}^c} \Big(\frac{\vmax}{\umax} +
      \frac{\diam(\domain)}{\vmax} \Big),  
    \end{split}
  \end{equation}
  where $\diam(\domain)$ is the length of the largest segment lying
  completely inside $\domain$. From the well known upper bound
  \cite{JMS:90} on the tour length of optimal ETSP, there exists a constant
  $\beta(\domain)$ such that $\sum_{i \in \mathcal{I}} \delta_i \leq
  \sum_{i=1}^n \delta_i \leq \beta(\domain) n^{1-\frac{1}{d}}$. Hence an
  upper bound on the term $\sum_{i \in \mathcal{I}} \sqrt{\delta_i}$ in
  eqn. \eqref{eq:first-expr-for-SOTSP} can be obtained by solving the
  following optimization problem:
  \begin{equation*}
    \begin{split}
      \maximize \quad & \sum_{i \in \mathcal{I}} \sqrt{\delta_i}, \qquad
      \subj \quad 
\sum_{i \in \mathcal{I}} \delta_i \leq \beta(\domain)
      n^{1-\frac{1}{d}}.
    \end{split}
  \end{equation*}
By employing the method of Lagrange multipliers, one can see that the
maximum is achieved when $\delta_i = \beta(\domain)
\frac{n^{1-\frac{1}{d}}}{n_{\mathcal{I}}} \quad \forall i \in \mathcal{I}$.
Hence $\sum_{i \in \mathcal{I}} \sqrt{\delta_i} \leq \sqrt{\beta(\domain)}
\sqrt{n_{\mathcal{I}} n^{1-\frac{1}{d}}}$. Substituting this in eqn.
\eqref{eq:first-expr-for-SOTSP}, we get that
\begin{equation}
  \label{eq:final-expr-for-SOTSP}
  \SOTSP{\pointset} \leq \frac{2 \sqrt{\beta(\domain)}}{\sqrt{\umax}}
  \sqrt{n_{\mathcal{I}}} n^{\frac{1}{2}-\frac{1}{2d}} + n_{\mathcal{I}^c}
  \Big(\frac{\vmax}{\umax} + \frac{\diam(\domain)}{\vmax} \Big).
\end{equation}
However, $n_{\mathcal{I}} \leq n$ and Lemma~\ref{lem:num-of-dist-pts-bound}
implies that $n_{\mathcal{I}^c}$ belongs to $O(n^{1-\frac{1}{d}})$.
Incorporating these facts into eqn. \eqref{eq:final-expr-for-SOTSP}, one
arrives at the final result.
\end{proof}

The above theorem relies on the following key result.
\begin{lemma}
\label{lem:num-of-dist-pts-bound}
Given any point set $\pointset \in \setpointsets{n}$ in $\domain \subset
\real^d$, if $(q_1,q_2,\ldots,q_n,q_1)$ is the order of points for the
optimal ETSP tour over $\pointset$, then for any $\eta \in \real_+$, the
cardinality of the set $\setdef{q_i \in \pointset}{\norm{q_i-q_{i+1}} >
  \eta}$ belongs to $O(n^{1-\frac{1}{d}})$.
\end{lemma}
\begin{proof}
  By contradiction, assume there exists $\tilde{\eta} \in \real_+$ such
  that the cardinality of $\setdef{p_i \in \pointset}{\norm{q_i-q_{i+1}} >
    \tilde{\eta}}$ belongs to $\Omega(n^{1-\frac{1}{d}+\epsilon})$ for some
  $\epsilon > 0$. This implies that $\ETSP{P}$ belongs to $\tilde{\eta}
  \times \Omega(n^{1-\frac{1}{d}+\epsilon}) =
  \Omega(n^{1-\frac{1}{d}+\epsilon})$.  However, we know from~\cite{JMS:90}
  that $\ETSP{P}\in O(n^{1-\frac{1}{d}})$.
\end{proof}

\section{The stochastic $\DoubIntTSP$}
\label{sec:stochastic-SOTSP}
The results in the previous section showed that based on a simple strategy, the STOP-GO-STOP strategy, we are already guaranteed to have sublinear cost for the $\DoubIntTSP$ when the point sets are considered on an individual basis. However, it is reasonable to argue that
there might be better algorithms when one is concerned with \emph{average} performance. In particular, one can expect that when
$n$ target points are stochastically generated in $\domain$ according to a
uniform probability distribution function, the cost of $\DoubIntTSP$ should be lower than the one given by the STOP-GO-STOP strategy. We shall refer to the problem of studying the average performance of $\DoubIntTSP$ over this class of point sets as \emph{stochastic} $\DoubIntTSP$. In this section, we present novel algorithms for stochastic $\DoubIntTSP$ and then establish bounds on their performances.

We make the following assumptions: in $\real^2$, $\domain$ is a rectangle of width
$W$ and height $H$ with $W\geq H$; in $\real^3$, $\domain$ is a rectangular box of
width $W$, height $H$ and depth $\depth$ with $W\geq H\geq \depth$.
Different choices for the shape of $\domain$ affect our conclusions only by
a constant.  The axes of the reference frame are parallel to the sides of
$\domain$.  The points $P = (p_1,\ldots,p_n)$ are randomly generated
according to a uniform distribution in $\domain$.

\subsection{Lower bounds}
First we provide lower bounds on the expected length of stochastic $\DoubIntTSP$ for the 2 and 3 dimensional case.
\begin{theorem} \textit{(Lower bounds on stochastic $\DoubIntTSP$)}
\label{theorem:lower-bounds}
For all $\vmax>0$ and $\umax>0$, the expected cost of a stochastic $\DoubIntTSP$ visiting a set of $n$ uniformly-randomly-generated points satisfies the following inequalities:
  \begin{eqnarray*}
    \lim_{n\to+\infty} \frac{\E[\SOTSP{\pointset \subset \domain \subset \real^2}]}{n^{2/3}} \ge
    \frac{3}{4} \Big(\frac{6 \width \height}{\vmax \umax}\Big)^{1/3} \quad \text{and} \\
    \lim_{n\to+\infty} \frac{\E[\SOTSP{\pointset \subset \domain \subset \real^3}]}{n^{4/5}} \ge
    \frac{5}{6} \Big(\frac{20 \width \height \depth}{\pi \vmax \umax^2}\Big)^{1/5}.
  \end{eqnarray*}
\end{theorem} 

\begin{proof}
We first prove the first inequality.
Choose a random point $q_i \in \pointset$ as the initial position and $v_i$ as the initial speed of the
  vehicle on the tour, and choose the heading randomly. We would like
  to compute a bound on the expected time to the closest next point in
  the tour; let us call such a time $t^*$.
To this purpose, consider the set $R_\mathrm{t}$ of points that are
  reachable by a second order vehicle within time $t$ .
  It can be verified that the area of such a set can be bounded, as
  $t\to0^+$, by
  \begin{equation}
    \mathrm{Area}(R_t) \leq \frac{\umax v_i t^3}{6} + o(t^3) \leq \frac{\umax \vmax t^3}{6} + o(t^3).
  \end{equation}
Given time $t$, the probability that $t^* > t$
  is no less than the probability that there is no other target
  reachable within a time at most $t$; in other words,
  \begin{equation*}
    \mathrm{Pr}[t^* > t] \ge 1 - n \frac{\mathrm{Area}(R_t)}{\mathrm{Area}(\domain)}
    \ge 1 - n \frac{\umax \vmax t^3}{6 \width \height} - o (t^3).    
  \end{equation*}
 In terms of expectation, defining $c = \frac{n \umax \vmax}{6 \width \height}$,
  \begin{eqnarray*}
    \E[t^*] &=& \int_0^{+\infty} \mathrm{Pr}(t^*> \xi) \;
    d\xi\\ &\ge & \int_0^{+\infty} \max\left\{0, 1-\frac{n \umax \vmax}{6 \width \height}\xi^3 - o
    (\xi^3)\right\}\; d\xi\\ & \ge & \int_0^{c^{-1/3}} (1- c\xi^3) \; d \xi
    - n \int_0^{c^{-1/3}} o(\xi^3) \; d \xi\\ 
    &=& \frac{3}{4}
    \left(\frac{6 \width \height}{\vmax \umax n}\right)^{1/3} - o (n^{-1/3}).
  \end{eqnarray*}
The expected total tour time will be no smaller than $n$ times the
  expected shortest time between two points, i.e.,
  $$\E[\SOTSP{\pointset}{\vmax,\umax}{2}] \geq \frac{3}{4} \left(\frac{6 n^2 \width \height}{\vmax \umax}\right)^{1/3} - o (n^{2/3}).$$
  Dividing both sides by $n^{2/3}$ and
  taking the limit as $n\to+\infty$, we get the first result.

We now prove the second inequality.
Choose a random point $q_i \in \pointset$ as the initial position and $v_i$ as the initial speed of the
  vehicle on the tour, and choose the heading randomly. We would like
  to compute a bound on the expected time to the closest next point in
  the tour; let us call such a time $t^*$.
To this purpose, consider the set $R_\mathrm{t}$ of points that are
  reachable by a second order vehicle within time $t$ .
  It can be verified that the volume of such a set can be bounded, as
  $t\to0^+$, by
  \begin{equation}
    \mathrm{Volume}(R_t) \leq \frac{\pi \umax^2 v_i t^5}{20} + o(t^5) \leq  \frac{\pi \umax^2 \vmax t^5}{20} + o(t^5).
  \end{equation}
Given time $t$, the probability that $t^* > t$
  is no less than the probability that there is no other target
  reachable within a time at most $t$; in other words,
  \begin{equation*}
    \mathrm{Pr}[t^* > t] \ge 1 - n \frac{\mathrm{Volume}(R_t)}{\mathrm{Volume}(\domain)}
    \ge 1 - n \frac{\pi \umax^2 \vmax t^5}{20 \width \height \depth} - o (t^5).    
  \end{equation*}
In terms of expectation, defining $c = \frac{n \pi \umax^2 \vmax}{20 \width \height \depth}$,
  \begin{eqnarray*}
    \E[t^*] &=& \int_0^{+\infty} \mathrm{Pr}(t^*> \xi) \;
    d\xi\\ &\ge & \int_0^{+\infty} \max\left\{0, 1-\frac{n \pi \umax^2 \vmax}{20 \width \height \depth} \xi^5 - o
    (\xi^5)\right\}\; d\xi\\ & \ge & \int_0^{c^{-1/5}} (1- c\xi^5) \; d \xi
    - n \int_0^{c^{-1/5}} o(\xi^5) \; d \xi\\ 
    &=& \frac{5}{6}
    \left(\frac{20 \width \height \depth}{\vmax \umax^2 n}\right)^{1/5} - o (n^{-1/5}).
  \end{eqnarray*}

The expected total tour time will be no smaller than $n$ times the
  expected shortest time between two points, i.e.,
  $$\E[\SOTSP{\pointset}{\vmax,\umax}{3}] \geq \frac{5}{6} \left(\frac{20 n^4 \width \height \depth}{\vmax \umax^2}\right)^{1/5} - o (n^{4/5}).$$
  Dividing both sides by $n^{4/5}$ and
  taking the limit as $n\to+\infty$, we get the second result.

\end{proof}

\subsection{Constructive upper bounds}
In this section, we first recall the \RecBeadTilingAlgo from our earlier work \cite{KS-EF-FB:06a} on \emph{Dubins} vehicle and use it to propose novel algorithms for the stochastic $\DoubIntTSP$: the \RecBeadTilingAlgo for $\real^2$ and \RecCylFillingAlgo for $\real^3$. The performances of these algorithms will be shown to be within a constant factor of the optimal with high probability. 

In \cite{KS-EF-FB:06a}, we studied stochastic versions of TSP for Dubins vehicle. Though conventionally Dubins vehicle is restricted to be a \emph{planar} vehicle, one can easily generalize the  model even for the three (and higher) dimensional case. Correspondingly, Dubins vehicle can be defined as a vehicle that is constrained to move with a constant speed along paths of bounded curvature, without reversing direction. Accordingly, a \emph{feasible curve for Dubins vehicle} or a \emph{Dubins path} is defined as a curve that is twice differentiable almost everywhere, and such that the magnitude of its curvature is bounded above
by $1/\rho$, where $\rho>0$ is the minimum turn radius. Based on this, one can immediately come up with the following analogy between feasible curves for Dubins vehicle and a double integrator.

\begin{lemma}\textit{(Trajectories of Dubins and double integrators)}
  \label{lem:feasible-curves}
  A feasible curve for Dubins vehicle with minimum turn radius $\rho>0$ is
  a feasible curve for a double integrator (modeled in
  equation~\eqref{eq:second-order-model}) moving with a constant speed
  $\sqrt{\rho \umax}$. Conversely, a feasible curve for a double integrator
  moving with a constant speed $s \leq \vmax$ is a feasible curve for Dubins
  vehicle with minimum turn radius $\frac{s^2}{\umax}$.
\end{lemma}\smallskip

In \cite{KS-EF-FB:06h-tmp}, we proposed a novel algorithm, the
\RecBeadTilingAlgo for the stochastic version of the Dubins TSP (DTSP) in
$\real^2$; we showed that this algorithm performed within a constant factor
of the optimal with high probability. In this paper, taking inspiration
from those ideas, we propose algorithms to compute feasible curves for a
double integrator moving with a constant speed.
%
Note that moving at the maximum speed $\vmax$ is not necessarily the best
strategy since it restricts the maneuvering capability of the vehicle.
Nonetheless, this strategy leads to efficient algorithms.
In what follows we assume that the double integrator is moving with some
constant speed $s \leq \vmax$. Next, we proceed towards devising strategies
which perform within a constant factor of the optimal for stochastic
$\DoubIntTSP$ in $\real^2$ as well as $\real^3$, both with high
probability.

\subsubsection{The basic geometric construction}
\label{sec:basic}
\begin{figure}[htb]
  \centerline{\includegraphics[width=0.5\columnwidth]{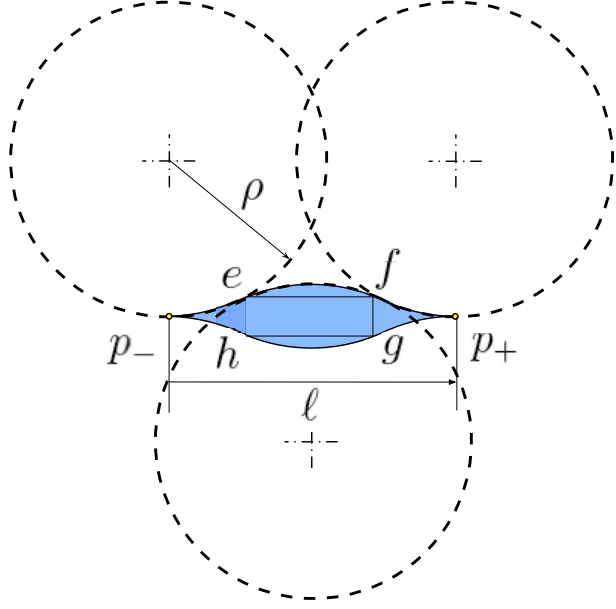}}
  \caption{Construction of the ``bead'' $\tile(\ell)$. The figure shows how the
    upper half of the boundary is constructed, the bottom half is
    symmetric.The figure shows the rectangle $efgh$ which is used to
    construct the "cylinder" $\cyl(\ell)$.}
  \label{fig:tile}
\end{figure}

Here we define useful geometric objects and study their properties.  Given
the constant speed $s$ for the double integrator let
$\rho=\frac{s^2}{\umax}$; from Lemma~\ref{lem:feasible-curves} this
constant corresponds to the minimum turning radius of the \emph{analogous}
Dubins vehicle. Consider two points $p_-$ and $p_+$ on the plane, with
$\ell = \|p_+-p_-\|_2 \le 4\rho$, and construct the bead $\tile(\ell)$ as
detailed in Figure~\ref{fig:tile}.

Associated with the bead is also the rectangle $efgh$. Rotating this
rectangle about the line passing through $p_-$ and $p_+$ gives rise to a
cylinder $\cyl(\ell)$. The regions $\tile(\ell)$ and $\cyl(\ell)$ enjoy the
following asymptotic properties as $(l/\rho)\to{0^+}$:
\begin{enumerate}
\item [(P1)] The maximum ``thickness'' of $\tile(\ell)$ is equal to
  \begin{equation*}
    w(\ell) = 4 \rho
    \left(1-\sqrt{1-\frac{\ell^2}{16\rho^2}}\right)=\frac{\ell^2}{8\rho}+
    \rho \cdot o\left(\frac{\ell^3}{\rho^3}\right) .
  \end{equation*}
The radius of cross-section of $\cyl(\ell)$ is $w(\ell)/4$ and the length of $\cyl(\ell)$ is $\ell$.

\item [(P2)] The area of $\tile(\ell)$ is equal to
  \begin{equation*}
    \Area(\tile(\ell)) = \frac{\ell w(\ell)}{2} = \frac{\ell^3}{16\rho} +
    \rho^2 \cdot o\left(\frac{\ell^4}{\rho^4}\right).
  \end{equation*}
  The volume of $\cyl(\ell)$ is equal to
  \begin{equation*}
  \Volume[\cyl(\ell)] = \pi \Big(\frac{w(\ell)}{4}\Big)^2 \frac{\ell}{2} = \frac{\pi \ell^5}{2048\rho^2} +
  \rho^3 \cdot o\left(\frac{\ell^6}{\rho^6}\right).
  \end{equation*}
  
\item [(P3)] For any $p \in \tile$, there is at least one feasible curve
  $\gamma_p$ through the points $\{p_-, p, p_+\}$, entirely contained
  within $\tile$.  The length of any such path is at most
  \begin{equation*}
    \Length(\gamma_p) \le 4\rho \arcsin\left(\frac{\ell}{4\rho}\right)
    = \ell + \rho \cdot o\left(\frac{\ell^3}{\rho^3}\right).
  \end{equation*}
  Analogously, for any $\tilde{p} \in \cyl$, there is at least one feasible
  curve $\gamma_{\tilde{p}}$ through the points $\{p_-, \tilde{p}, p_+\}$,
  entirely contained within the region obtained by rotating $\tile(\ell)$
  about the line passing through $p_-$ and $p_+$. The length of
  $\gamma_{\tilde{p}}$ satisfies the same upper bound as the one
  established for $\gamma_p$.

\end{enumerate}
The geometric shapes introduced above can be used to cover $\real^2$ and
$\real^3$ in an \emph{organized} way. The plane can be periodically
\emph{tiled}\footnote{A tiling of the plane is a collection of sets whose
  intersection has measure zero and whose union covers the plane.} by
identical copies of $\tile(\ell)$, for any $\ell\in]0,4\rho]$.  The
cylinder, however does not enjoy any such special property.  For our
purpose, we consider a particular covering of $\real^3$ by cylinders
described as follows.

\begin{figure}[htbp]
  \centerline{\includegraphics[width=0.5\linewidth]{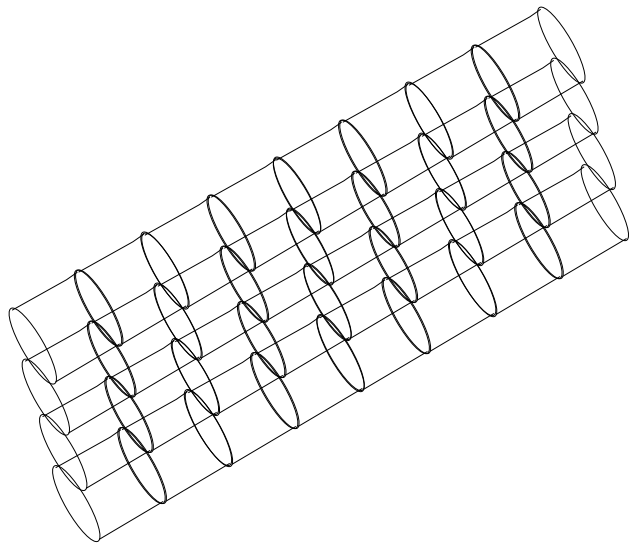}}
  \caption{A typical layer of cylinders formed by stacking rows of cylinders}
  \label{fig:layer-of-cylinders}
\end{figure}

\begin{figure}[htbp]
\begin{center}
  \includegraphics[width=0.42\linewidth]{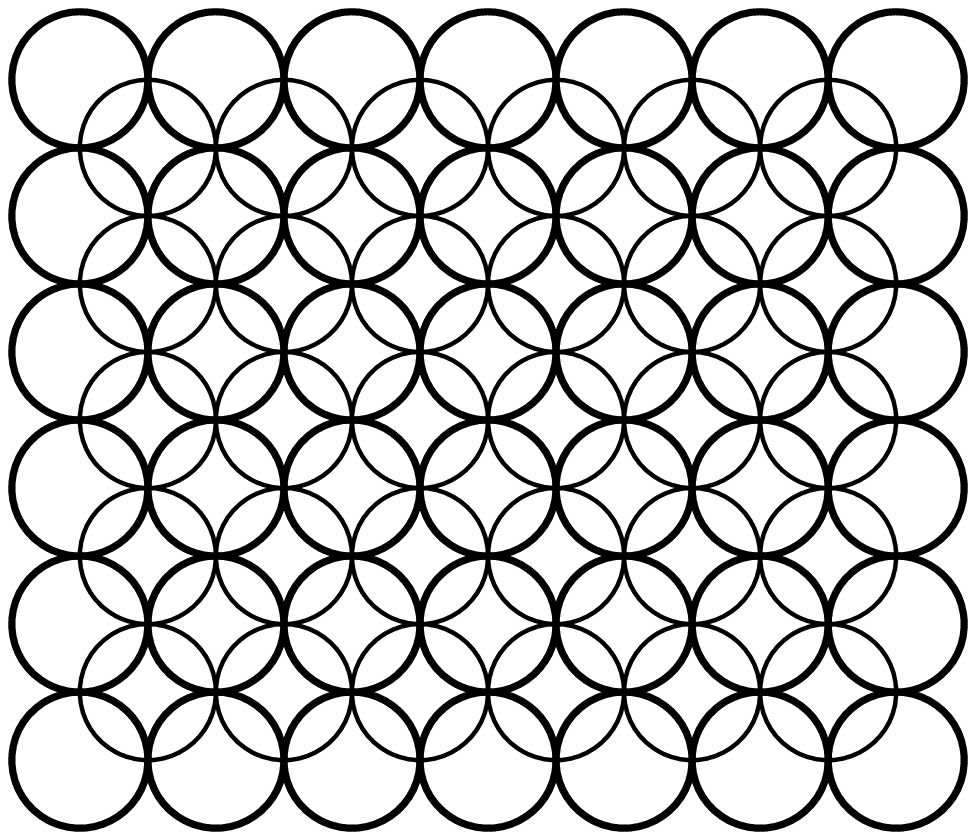}
  \includegraphics[width=0.42\linewidth]{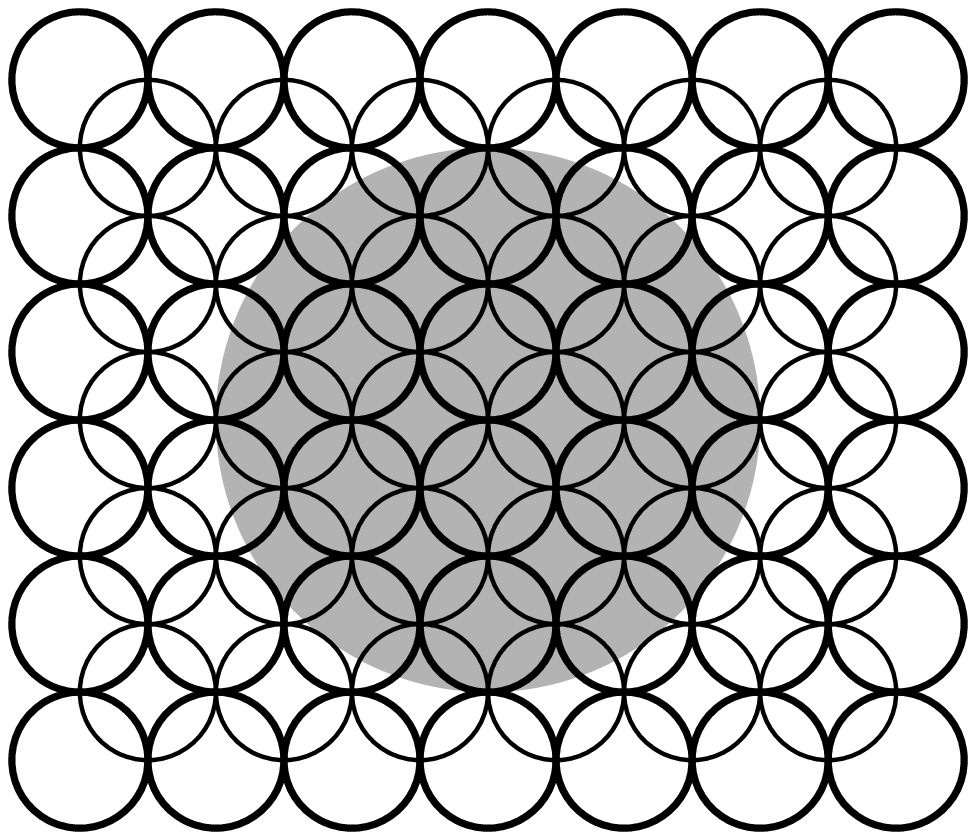}\\
  (a) \phantom{a big big b blank space} (b)
\end{center}
\caption{(a): Cross section of the arrangement of the layers of cylinders
  used for covering $\domain \subset \real^3$, (b): The relative position
  of the bigger cylinder relative to smaller ones of the prior phase during
  the phase transition.}
\label{fig:cross-section}  
\end{figure}

A \emph{row of cylinders} is formed by joining cylinders end to end along
their length. A layer of cylinders is formed by placing rows of cylinders
parallel and on top of each other as shown in
Figure~\ref{fig:layer-of-cylinders}. For covering $\real^3$, these layers
are arranged next to each other and with offsets as shown in
Figure~\ref{fig:cross-section}(a), where the cross section of this
arrangement is shown.  We refer to this construction as the
\emph{covering of $\real^3$}. 



\subsubsection{The 2D case: The \RecBeadTilingAlgo (\RecBTA)}
\begin{figure*}[htbp]
  \centerline{\includegraphics[width=0.95\linewidth]{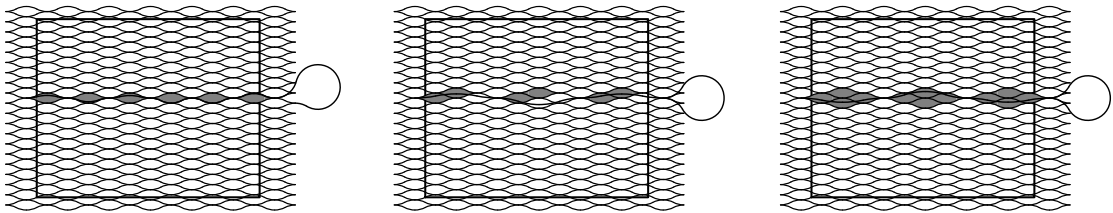}}
  \caption{Sketch of ``meta-beads" at successive phases in the recursive bead tiling algorithm.}
  \label{fig:sweep4}
\end{figure*}

Consider a tiling of the plane such that $\mathrm{Area}[\tile(\ell)] =
\mathrm{Area}[\domain \subset \real^2]/(2n) = WH/(2n)$; to obtain this
equality we assume $\ell$ to be a decreasing function of $n$ such that
$\ell(n) \leq 4\rho$. Furthermore, we assume the tiling is chosen to be
aligned with the sides of $\domain \subset \real^2$, see
Figure~\ref{fig:sweep4}.
 
The proposed algorithm consists of a sequence of phases; during each of
these phases, a feasible curve will be constructed that ``sweeps'' the set
$\domain$.  In the first phase, a feasible curve is constructed with the
following properties:
 \begin{enumerate}
 \item it visits all non-empty beads once,
 \item it visits all rows\footnote{A row is a maximal string of beads with
     non-empty intersection with~$\domain$.} in sequence top-to-down,
   alternating between left-to-right and right-to-left passes, and visiting
   all non-empty beads in a row,
 \item when visiting a non-empty bead, it services at least one target in
   it.
\end{enumerate}

In order to visit the outstanding targets, a new phase is initiated. In
this phase, instead of considering single beads, we will consider
``meta-beads'' composed of two beads each, as shown in
Figure~\ref{fig:sweep4}, and proceed in a similar way as the first phase,
i.e., a feasible curve is constructed with the following properties:
 \begin{enumerate}
  \item the curve visits all non-empty meta-beads once,
  \item it visits all (meta-bead) rows in sequence top-to-down,
    alternating between left-to-right and right-to-left passes, and
    visiting all non-empty meta-beads in a row,
  \item when visiting a non-empty meta-bead, it services at least one target in
    it.
\end{enumerate}

This process is iterated at most $\log_2 n+1$ times, and at each phase
meta-beads composed of two neighboring meta-beads from the previous phase
are considered; in other words, the meta-beads at the $i$-th phase are
composed of $2^{i-1}$ neighboring beads. After the last phase, the leftover
targets will be visited using, for example, a greedy strategy.
 
The following result is related to a similar result in
\cite{YA-AZB-ARK-EU:99}.

\begin{theorem}[Targets remaining after recursive phases]
  \label{theorem:RecBeadTilingAlgo}
  Let $\pointset\in\setpointsets{n}$ be uniformly randomly generated in
  $\domain \in \real^2$. The number of unvisited targets after the last
  recursive phase of the \RecBTA is less than $24 \log_2 n$ with high
  probability, i.e., with probability approaching one as $n\to+\infty$.
\end{theorem}\smallskip

\begin{proof}
  Associate a unique identifier to each bead, let $b(t)$ be the identifier
  of the bead in which the $t^{\text{th}}$ target is sampled, and let $h(t)
  \in \natural$ be the phase at which the $t^{\text{th}}$ target is
  visited.  Without loss of generality, assume that targets within a single
  bead are visited in the same order in which they are generated, i.e., if
  $b(t_1) = b(t_2)$ and $t_1 < t_2$, then $h(t_1) < h(t_2)$.  Let $v_i(t)$
  be the number of beads that contain unvisited targets at the inception of
  the $i^{\text{th}}$ phase, computed after the insertion of the
  $t^{\text{th}}$ target. Furthermore, let $m_i$ be the number of
  $i^{\text{th}}$ phase meta-beads (i.e., meta-beads containing $2^{i-1}$
  neighboring beads) with a non-empty intersection with $\domain$. Clearly,
  $v_i(t) \le v_i(n)$, $m_i \le 2m_{i+1}$, and $v_1(n) \le n \le m_1/2$
  with certainty.  The $t^{\text{th}}$ target will not be visited during
  the first phase if it is sampled in a bead that already contains other
  targets. In other words,
  \begin{equation*}
    \Pr\big[h(t) \ge 2|\; v_1(t)\big] = \frac{v_1(t)}{m_1} \le \frac{v_1(n)}{2n}
    \le \frac{1}{2}.     
  \end{equation*}
  Similarly, the $t^{\text{th}}$ target will not be visited during the
  $i^{\text{th}}$ phase if (i) it has not been visited before the
  $i^{\text{th}}$ pass, and (ii) it belongs to a meta-bead that already
  contains other targets not visited before the $i^{\text{th}}$ phase:
  \begin{align*}
    \Pr\big[h(t)  \ge i+1 &|\; (v_i(t-1), v_{i-1}(t-1), v_1(t-1))\big] \\
    &= \Pr\big[h(t) \ge i+1 |\;  h(t) \ge i, v_i(t-1)\big]  
     \; \cdot\; \Pr\big[h(t) \ge i|\; (v_{i-1}(t-1), \ldots, v_1(t-1))\big] \\
    &\le  \frac{v_i(t-1)}{m_i} \Pr[h(t) \ge i |\; (v_{i-1}(t-1), \ldots,
    v_1(t-1))] \\
    &= \prod_{j=1}^i \frac{v_j(t-1)}{m_j} \le \prod_{j=1}^i \frac{2^{j-1}
      v_j(n)}{2n} = \left(\frac{2^{\frac{i-3}{2}}}{n}\right)^i \prod_{j=1}^i
    v_j(n).
  \end{align*}
  
  Given a sequence $\{\beta_i\}_{i\in\natural}\subset\real_+$ and given a
  fixed $i \ge 1$, define a sequence of binary random variables
  \begin{equation*}
    Y_t =
    \begin{cases}
      1,\quad & \text{if}\;  h(t) \ge i+1 \mbox{ and } v_i(t-1) \le \beta_i n,\\
      0, &\text{otherwise.}
    \end{cases}
  \end{equation*}
  In other words, $Y_t=1$ if the $t^{\text{th}}$ target is not visited
  during the first $i$ phases even though the number of beads still
  containing unvisited targets at the inception of the $i^{\text{th}}$
  phase is less than $\beta_i n$.  Even though the random variable $Y_t$
  depends on the targets generated before the $t^{\text{th}}$ target, the
  probability that it takes the value 1 is bounded by
  \begin{equation*}
    \Pr[Y_t = 1 |\; b(1), b(2), \ldots, b(t-1)] \le 2^{\frac{i(i-3)}{2}}
    \prod_{j=1}^i \beta_j =: q_i,
  \end{equation*}
  regardless of the actual values of $b(1), \ldots, b(t-1)$.  It is
  known~\cite{YA-AZB-ARK-EU:99} that if the random variables $Y_t$ satisfy
  such a condition, the sum $\sum_t Y_t$ is stochastically dominated by a
  binomially distributed random variable, namely,
  \begin{equation*}
    \Pr\left[\sum_{t=1}^n Y_t  > k\right] \le \Pr[B(n,q_i) > k].
  \end{equation*}
  In particular,
  \begin{equation}
    \label{eq:chernoff}
    \Pr\left[\sum_{t=1}^n Y_t > 2nq_i\right] \le \Pr[B(n,q_i) > 2 np_i] <
    2^{-nq_i/3}, 
  \end{equation}
  where the last inequality follows from Chernoff's Bound~\cite{RM-PR:95}.
  Now, it is convenient to define $\{\beta_i\}_{i\in\natural}$ by
  \begin{equation*}
    \beta_1 = 1,\quad
    \beta_{i+1} =  2 q_i = 2^{\frac{i(i-3)}{2}+1}
    \prod_{j=1}^i \beta_j = 2^{i-2} \; \beta_i^2,
  \end{equation*}
  which leads to $\beta_i = 2^{1-i}$. In turn, this implies that
  equation~\eqref{eq:chernoff} can be rewritten as
  \begin{equation*}
    \Pr\left[\sum_{t=1}^n Y_t > \beta_{i+1}n\right]  
    < 2^{-\beta_{i+1}n/6}=2^{-\frac{n}{3 \cdot  2^{i}}}, 
  \end{equation*}
  which is less than $1/n^2$ for $i \le i^*(n) := \lfloor \log_2 n - \log_2
  \log_2 n - \log_2 6 \rfloor \le \log_2 n$.  Note that $\beta_{i} \le 12
  \; \frac{\log_2 n}{n}$, for all $i > i^*(n)$.
  
  Let $\mathcal{E}_i$ be the event that $v_i(n) \le \beta_i n$.  Note that
  if $\mathcal{E}_i$ is true, then $v_{i+1}(n) \le \sum_{t=1}^{n} Y_t$: the
  right hand side represents the number of targets that will be visited
  after the $i^{\text{th}}$ phase, whereas the left hand side counts the
  number of beads containing such targets. We have, for all $i \le i^*(n)$:
  \begin{equation*}
    \Pr\Big[ v_{i+1} > \beta_{i+1}n |\; \mathcal{E}_i \Big] \cdot
    \Pr[\mathcal{E}_i]  \le \Pr\left[\sum_{t=1}^n Y_t > \beta_{i+1} n\right]
    \le \frac{1}{n^2},
  \end{equation*}
  that is, $ \displaystyle \Pr\left[\neg \mathcal{E}_{i+1} |\;
    \mathcal{E}_i \right] \le \frac{1}{n^2\; \Pr[\mathcal{E}_i]},$
  and thus (recall that $\mathcal{E}_1$ is true with certainty):
  \begin{equation*}
    \Pr\left[\neg \mathcal{E}_{i+1}\right]  \le \frac{1}{n^2} + \Pr[\neg
    \mathcal{E}_i] \le \frac{i}{n^2}.
  \end{equation*}
  In other words, for all $i \le i^*(n)$, $v_{i}(n) \le \beta_in$ with high
  probability.
  
  Let us now turn our attention to the phases such that $i>i^*(n)$. The
  total number of targets visited after the $(i^*)^{\text{th}}$ phase is
  dominated by a binomial variable $B(n, 12 \log_2 n/n)$; in particular,
  \begin{align*}
    \Pr \Big[ v_{i^*+1} > 24 \log_2 n |\; \mathcal{E}_{i^*}\Big] \cdot
    \Pr[\mathcal{E}_{i^*}] &\le \Pr \Big[ \sum_{t=1}^n Y_t > 24 \log_2 n\Big] \\
    &\le \Pr \big[B(n, 12 \log_2 n/n) > 24 \log_2 n\big] \le 2^{-12 \log_2
      n}.
  \end{align*}
  Dealing with conditioning as before, we obtain
  \begin{equation*}
    \Pr \left[v_{i^*+1} > 24 \log_2 n\right] \le \frac{1}{n^{12}} + \Pr[\neg
    \mathcal{E}_{i^*}] \le \frac{1}{n^{12}} + \frac{\log_2 n}{n^2}.
  \end{equation*}
  In other words, the number of targets that are left unvisited after the
  $(i^*)^{\text{th}}$ phase is bounded by a logarithmic function of $n$
  with high probability.
\end{proof}

In summary, Theorem~\ref{theorem:RecBeadTilingAlgo} says that after a
sufficiently large number of phases, almost all targets will be visited,
with high probability.  The second key point is to recognize that (i) the
length of the first phase is of order $n^{2/3}$ and (ii) the length of each
phase is decreasing at such a rate that the sum of the lengths of the
$\lceil\log_2 n\rceil$ recursive phases remains bounded and proportional to
the length of the first phase.  (Since we are considering the asymptotic
case in which the number of targets is very large, the length of the beads
will be very small; in the remainder of this section we will tacitly
consider the asymptotic behavior as $\ell/\rho \to 0^+$.)

\begin{lemma}[Path length for the first phase]
  \label{lemma:first_phase} 
  Consider a tiling of the plane with beads of length $\ell$. For any $\rho
  > 0$ and for any set of target points, the length $L_1$ of a path
  visiting once and only once each bead with a non-empty intersection with
  a rectangle $\domain$ of width $W$ and length $H$ satisfies
  \begin{equation*}
    L_1 \le \frac{16 \rho WH}{\ell^2} \left( 1+ \frac{7}{3} \pi \frac{\rho}{W}
    \right)+ \rho \cdot o\left( \frac{\rho}{\ell}\right).
  \end{equation*}
\end{lemma}\smallskip
\begin{proof}
  A path visiting each bead once can be constructed by a sequence of
  passes, during which all beads in a row are visited in a left-to-right or
  right-to-left order. In each row, there are at most $\lceil W/\ell\rceil
  +1$ beads with a non-empty intersection with $\domain$. Hence, the cost
  of each pass is at most:
  \begin{equation*}
  L^\mathrm{pass}_1 \le W + 2 \ell + \rho \cdot
  o\left(\frac{\ell^2}{\rho^2}\right).
  \end{equation*} 
  
  Two passes are connected by a U-turn maneuver, in which the direction of
  travel is reversed, and the path moves to the next row, at distance equal
  to one half the width of a bead. The length of the shortest path to
  reverse the heading of the vehicle with co-located initial and final
  points is $(7/3) \pi \rho$, the length of the U-turn satisfies
  \begin{equation*}
    L^\mathrm{U-turn}_1 \le \frac{7}{3}\pi \rho + \frac{1}{2}w(\ell) \le
    \frac{7}{3} \pi \rho + \frac{\ell^2}{16 \rho} + \rho \cdot o\left(
      \frac{\ell^3}{\rho^3}\right).    
  \end{equation*}  
  The total number of passes, i.e., the total number of rows of beads with
  non-empty intersection with $\domain$, satisfies 
  \begin{equation*}
    N^\mathrm{pass}_1 \le \left\lceil \frac{2H}{w(\ell)} \right\rceil + 1
    \le \frac{16\rho H}{\ell^2} + 2 +
    o\left(\frac{\rho}{\ell}\right).
  \end{equation*}  
  A simple upper bound on the cost of closing the tour is given by 
  \begin{equation*}
    L^\mathrm{close}_1 \le (W + 2 \ell) + (H + 2 w(\ell)) + 2\pi \rho = W +
    H + 2\pi\rho + 2 \ell + \rho \cdot o(\ell/\rho).
  \end{equation*}  
  In summary, the total length of the path followed during the first phase
  is
  \begin{align*}
    L_1 & \le N^\mathrm{pass}_1 \left(L^\mathrm{pass}_1 +
      L^\mathrm{U-turn}_1\right) + L^\mathrm{close}
    \\
    &\le \left(\frac{16\rho H}{\ell^2} + 2 +
      o\left(\frac{\rho}{\ell}\right)\right) \left( W + 2 \ell +
      \frac{7}{3} \pi \rho + \frac{\ell^2}{16 \rho}+\rho \cdot o\left(
        \frac{\ell^2}{\rho^2}\right) \right) + W + H + 2 \pi \rho + 2 \ell
    + \rho \cdot o(\ell/\rho)
    \\
    &\le \frac{16 \rho WH}{\ell^2} \left( 1+ \frac{7}{3} \pi \frac{\rho}{W}
    \right)+ \rho \cdot o \left( \frac{\rho}{\ell}\right).
\end{align*}
\end{proof}

Based on the calculation for the first phase, we can estimate the length of the paths in
generic phases of the algorithm. Since the total number of phases in the
algorithm depends on the number of targets $n$, as does the length of the
beads $\ell$, we will retain explicitly the dependency on the phase number.
 
\begin{lemma}[Path length at odd-numbered phases]
  \label{lemma:odd}
  Consider a tiling of the plane with beads of length $\ell$. For any $\rho
  > 0$ and for any set of target points, the length $L_{2j-1}$ of a path
  visiting once and only once each meta-bead with a non-empty intersection
  with a rectangle $\domain$ of width $W$ and length $H$ at phase number
  $(2j-1)$, $j\in\natural$ satisfies
 \begin{equation*}
    L_{2j-1} 
      \leq 2^{5-j} \left[\frac{\rho W H}{\ell^2} \left(1 +
        \frac{7}{3}\frac{\pi \rho}{W} \right) + \rho \cdot
      o\left(\frac{\rho}{\ell} \right) \right] + 32 \frac{\rho H}{\ell} 
      + \rho \cdot o\left(\frac{\rho}{\ell} \right) 
      + 2^j \left[3 \ell+ \rho \cdot  o\left(\frac{\ell}{\rho} \right) \right]. 
  \end{equation*}
\end{lemma}\smallskip
\begin{proof}
  During odd-numbered phases, the number of beads in a meta-bead is a
  perfect square and the considerations made in the proof of
  Lemma~\ref{lemma:first_phase} can be readily adapted.  The length of each
  pass satisfies
  \begin{equation*}
    L^\mathrm{pass}_{2j-1} \le \left(W + 2^{j} \ell\right) \left[ 1 +
    o\left(\frac{\ell}{\rho}\right)\right].      
  \end{equation*}
  The length of each U-turn maneuver is bounded as
  \begin{equation*}
    L^\mathrm{U-turn}_{2j-1} \le \frac{7}{3}\pi \rho + 2^{j-2}w(\ell) \le
    \frac{7}{3}\pi \rho + 2^{j-2}\;\left[ \frac{\ell^2}{8\rho} +  \rho \cdot
    o\left(\frac{\ell^3}{\rho^3}\right)\right],
  \end{equation*}  
  from which 
  \begin{equation*}
    L^\mathrm{pass}_{2j-1} + L^\mathrm{U-turn}_{2j-1} = W+ \frac{7}{3}\pi
    \rho + o \left(\frac{\ell}{\rho}\right) + 2^j \left[ \ell + \rho \cdot o
      \left(\frac{\ell}{\rho}\right)\right]. 
  \end{equation*}
    
  The number of passes satisfies:
  \begin{equation*}
    N^\mathrm{pass}_{2j-1} \le 2^{5-j} \left[\frac{\rho
    H}{\ell^2}+o\left(\frac{\rho}{\ell}\right)\right] + 2.    
  \end{equation*}
  Finally, the cost of closing the tour is bounded by
  \begin{equation*}
    L^\mathrm{close}_{2j-1} \le W + H +  2\pi \rho + 2^{j} \left[\ell +
      \rho \cdot o(\ell/\rho)\right].
  \end{equation*}
  Therefore, a bound on the total length of the path is 
  \begin{multline*}
    L_{2j-1} = N^\mathrm{pass}_{2j-1} (
    L^\mathrm{pass}_{2j-1}+L^\mathrm{U-turn}_{2j-1}) +
    L^\mathrm{close}_{2j-1}\\
    \leq 2^{5-j} \left[\frac{\rho W H}{\ell^2} \left(1 +
        \frac{7}{3}\frac{\pi \rho}{W} \right) + \rho \cdot
      o\left(\frac{\rho}{\ell} \right) \right] + 32 \frac{\rho H}{\ell} 
      + \rho \cdot o\left(\frac{\rho}{\ell} \right) 
      + 2^j \left[3 \ell+ \rho \cdot  o\left(\frac{\ell}{\rho} \right) \right].
  \end{multline*}
\end{proof}

\begin{lemma}[Path length at even-numbered phases]
  \label{lemma:even} 
  Consider a tiling of the plane with beads of length $\ell$. For any $\rho
  > 0$, a rectangle $\domain$ of width $W$ and length $H$ and any set of
  target points, paths in each phase of the $\BeadTilingAlgo$ can be chosen
  such that $L_{2j} \le 2 L_{2j+1}$, for all $j \in \natural$.
\end{lemma}\smallskip
\begin{proof}
  Consider a generic meta-bead $B_{2j+1}$ traversed in the
  $(2j+1)^\mathrm{th}$ phase, and let $l_3$ be the length of the path
  segment within $B_{2j+1}$. The same meta-bead is traversed at most twice
  during the $(2j)^\mathrm{th}$ phase; let $l_1$, $l_2$ be the lengths of
  the two path segments of the $(2j)^\mathrm{th}$ phase within $B_{2j+1}$.
  By convention, for $i\in\{1,2,3\}$, we let $l_i=0$ if the $i^\mathrm{th}$
  path does not intersect $B_{2j+1}$. Without loss of generality, the order
  of target points can be chosen in such a way that $l_1 \le l_2 \le l_3$,
  and hence $l_1+l_2 \le 2 l_3$. Repeating the same argument for all
  non-empty meta-beads, we prove the claim.
\end{proof}

Finally, we can summarize these intermediate bounds into the main result of
this section. We let $\LenRBTA{\pointset}{\rho}$ denote the length of the
path computed by the \RecBeadTilingAlgo for a point set~$\pointset$.
\begin{theorem}[Path length for the \RecBeadTilingAlgo]
  \label{theorem:total-path-upper-bound}
  Let $P\in\setpointsets{n}$ be uniformly randomly generated in the
  rectangle of width $W$ and height $H$. For any $\rho>0$, with high
  probability
  \begin{equation*}
    \lim_{n\to+\infty} \frac{\DTSP{P}{\rho}}{n^{2/3}} \,\le\,
    \lim_{n\to+\infty} \frac{ \LenRBTA{\pointset}{\rho} }{n^{2/3}} 
    \, \le \, 24 \sqrt[3]{\rho W H} \left( 1+ \frac{7}{3} \pi \frac{\rho}{W}
    \right).    
  \end{equation*}
\end{theorem}\smallskip
\begin{proof}
  For simplicity we let $\LenRBTA{\pointset}{\rho}=L_\mathrm{RBTA}$.
  Clearly, $L_\mathrm{RBTA} = L'_\mathrm{RBTA} + L''_\mathrm{RBTA}$, where
  $L'_\mathrm{RBTA}$ is the path length of the first $\lceil\log_2n\rceil$
  phases of the $\RecBeadTilingAlgo$ and $L''_\mathrm{BTA}$ is the length
  of the path required to visit all remaining targets.  An immediate
  consequence of Lemma~\ref{lemma:even}, is that
  \begin{equation*}
    L'_\mathrm{RBTA} = \sum_{i=1}^{\lceil\log_2(n)\rceil} L_i
    \leq 3\sum_{j=1}^{\left\lceil{\log_2(n)}/{2}\right\rceil}
    L_{2j-1}.
  \end{equation*}
  The summation on the right hand side of this equation can be expanded
  using Lemma~\ref{lemma:odd}, yielding
  \begin{multline*}
    L'_\mathrm{RBTA} \le 3 \left\{ \left[\frac{\rho W H}{\ell^2} \left(1 +
      \frac{7}{3}\frac{\pi \rho}{W} \right) + \rho \cdot
      o\left(\frac{\rho^2}{\ell^2} \right) \right] 
      \sum_{j=1}^{\left\lceil\log_2(n)/2\right\rceil}2^{5-j} \right.\\ \left. 
      + \left( 32 \frac{\rho H}{\ell} + \rho \cdot
      o\left(\frac{\rho}{\ell}\right) \right) \left\lceil \frac{\log_2n}{2}
      \right\rceil + 
      \left[ 3 \ell + \rho \cdot
        o(\ell/\rho)\right]\sum_{j=1}^{\left\lceil\log_2(n)/2\right\rceil}2^j\right\}.
  \end{multline*}
  Since $\sum_{j=1}^k 2^{-j} \le \sum_{j=1}^{+\infty} 2^{-j} = 1,$ and
  $\sum_{j=1}^k 2^j = 2^{k+1}-2 \le 2^{k+1},$ the previous equation can be
  simplified to
  \begin{multline*}
    L'_\mathrm{RBTA} \le 3 \left\{ 32 \left[\frac{\rho W H}{\ell^2} \left(1
          + \frac{7}{3}\frac{\pi \rho}{W} \right) + \rho \cdot
        o\left(\frac{\rho}{\ell} \right) \right] \right.
    \\
    \left.  + \left(32 \frac{\rho H}{\ell} + \rho \cdot o\left(
          \frac{\ell}{\rho} \right) \right) \left\lceil \frac{\log_2n}{2}
      \right\rceil + \left[3 \ell + \rho \cdot o(\ell/\rho)\right]\cdot (4
      \sqrt{n}) \right\}.
  \end{multline*}  
  Recalling that $\ell = 2(\rho WH/n)^{1/3} + o(n^{-1/3})$ for large $n$,
  the above can be rewritten as
  \begin{equation*}
    L'_\mathrm{RBTA} \le 24 \sqrt[3]{\rho W H n^2}
    \left( 1+ \frac{7}{3} \pi \frac{\rho}{W} \right)+ o(n^{2/3}).
  \end{equation*}
  Now it suffices to show that $L''_\mathrm{RBTA}$ is negligible with
  respect to $L'_\mathrm{RBTA}$ for large $n$ with high probability.  From
  Theorem~\ref{theorem:RecBeadTilingAlgo}, we know that with high
  probability there will be at most $24 \log_2 n$ unvisited targets after
  the $\lceil\log_2n\rceil$ recursive phases. From
  \cite{KS-EF-FB:04l} we know that, with high probability, the
  length of a \AltAlgo tour through these points satisfies
  \begin{equation*}
    L''_\mathrm{RBTA} \le \kappa \lceil 12 \log_2 n \rceil \pi \rho+ o(\log_2
    n).   
  \end{equation*}
\end{proof}

In order to obtain an upper bound on the $\SOTSP{\pointset}$, we derive the
expression for time taken, $\mathcal{T}_\mathrm{RecBTA}$, by the \RecBTA to
execute the path of length $\LenRBTA{\pointset}{\rho}$ and then optimize it
with respect to $\rho$. Based on this calculation, we get the following
result.

\begin{theorem}\textit{(Upper bound on the total time in $\real^2$)}
  \label{theorem:total-time-upper-bound-2D}
  Let $P\in\setpointsets{n}$ be uniformly randomly generated in the
  rectangle of width $W$ and height $H$. For any double integrator
  \eqref{eq:second-order-model}, with high probability
  \begin{equation*}
    \lim_{n\to+\infty} \frac{\mathcal{T}_\mathrm{RecBTA}}{n^{2/3}} 
    \, \le \, 24 \left(\frac{WH}{\vmax \umax}\right)^{1/3} \left( 1+ \frac{7\pi\vmax^2}{3W}
    \right). 
  \end{equation*}
\end{theorem}\smallskip

\begin{remark}
  Theorems~\ref{theorem:lower-bounds}
  and~\ref{theorem:total-time-upper-bound-2D} imply that, with high
  probability, the \RecBTA is a
  $\frac{32}{\sqrt[3]{6}}\left(1+\frac{7\pi\vmax^2}{3\umax W}
  \right)$-factor approximation (with respect to $n$) to the optimal stochastic DITSP
  in $\real^2$ and that $\E[\SOTSP{\pointset \subset \domain \subset \real^2}]$
  belongs to $\Theta(n^{2/3})$.
\end{remark}
\subsubsection{The 3D case: The \RecCylFillingAlgo (\RecCCA)}
Consider a covering of $\domain \in \real^3$ by cylinders such that
$\mathrm{Volume}[\cyl(\ell)]= \mathrm{Volume}[\domain \subset \real^3]/(4n)
= WH\depth/(4n)$ (Again implying that $n$ is sufficiently large).
Furthermore, the covering is chosen in such a way that it is aligned with
the sides of $\domain \subset \real^3$.

The proposed algorithm will consist of a sequence of phases; each phase
will consist of five sub-phases, all similar in nature. For the first
sub-phase of the first phase, a feasible curve is constructed with the
following properties:
 \begin{enumerate}
  \item it visits all non-empty cylinders once,
  \item it visits all rows of cylinders in a layer in sequence top-to-down in a layer,
    alternating between left-to-right and right-to-left passes, and
    visiting all non-empty cylinders in a row,
  \item it visits all layers in sequence from one end of the region to the other,
  \item when visiting a non-empty cylinder, it services at least one target in
    it.
\end{enumerate}

\begin{figure*}[htbp]
  \centerline{\includegraphics[width=0.95\linewidth]{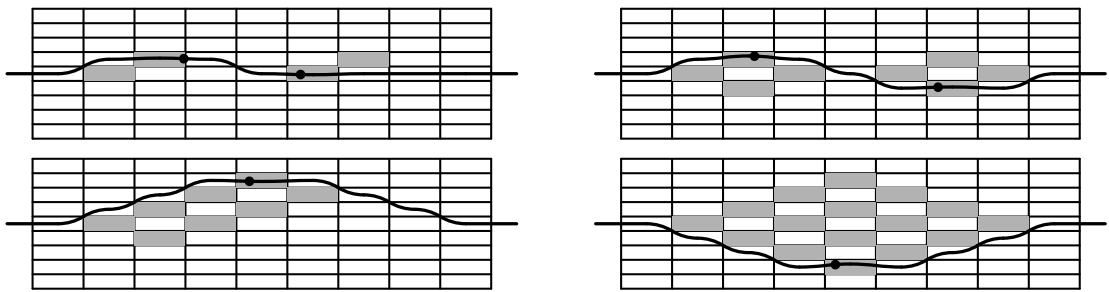}}
  \caption{Starting from top left in the left-to-right, top-to bottom
    direction, sketch of projection of ``meta-cylinders'' on the
    corresponding side of $\domain \subset \real^3$ at second, third,
    fourth and fifth sub-phases of a phase in the recursive cylinder
    covering algorithm.} 
  \label{fig:allsubphases}
\end{figure*}

In subsequent sub-phases, instead of considering single cylinders, we will
consider ``meta-cylinders'' composed of $2$, $4$, $8$ and $16$ beads each
for the remaining four sub-phases, as shown in
Figure~\ref{fig:allsubphases}, and proceed in a similar way as the first
sub-phase, i.e., a feasible curve is constructed with the following
properties:
 \begin{enumerate}
  \item the curve visits all non-empty meta-cylinders once,
  \item it visits all (meta-cylinder) rows in sequence top-to-down in a
    (meta-cylinder) layer, alternating between left-to-right and
    right-to-left passes, and visiting all non-empty meta-cylinders in a
    row,
  \item it visits all (meta-cylinder) layers in sequence from one end of
    the region to the other,
  \item when visiting a non-empty meta-cylinder, it services at least one
    target in it.
\end{enumerate}

A meta-cylinder at the end of the fifth sub-phase, and hence at the end of
the first phase will consist of 16 nearby cylinders. After this phase, the
transitioning to the next phase will involve enlarging the cylinder to $32$
times its current size by increasing the radius of its cross section by a
factor of $4$ and doubling its length as outlined in
Figure~\ref{fig:cross-section}(b). It is easy to see that this bigger
cylinder will contain the union of $32$ nearby smaller cylinders. In other
words, we are forming the object $\cyl(2\ell)$ using a conglomeration of $32$
$\cyl(\ell)$ objects. This whole process is repeated at most $\log_2{n}+2$ times. After the last phase, the leftover
targets will be visited using, for example, a greedy strategy.

We have the following results, which are similar to the one for the \RecBeadTilingAlgo.

\begin{theorem}[Targets remaining after recursive phases]
  \label{theorem:RecCylFillingAlgo}
  Let $\pointset\in\setpointsets{n}$ be uniformly randomly generated in
  $\domain \subset \real^3$. The number of unvisited targets after the last recursive phase
  of the \RecCylFillingAlgo over $\pointset$ is less than $24 \log_2 n$
  with high probability, i.e., with probability approaching one as
  $n\to+\infty$.
\end{theorem}\smallskip

\begin{lemma}[Path length for the first sub phase]
  \label{lemma:first_sub_phase} 
  Consider a covering of the space with cylinders $\cyl(\ell)$. For any $\rho
  > 0$ and for any set of target points, the length $L_I$ of a path executing the first sub-phase of the \RecCylFillingAlgo in a rectangular box $\domain$ of width $\width$, height $\height$ and depth $\depth$ satisfies
  \begin{equation*}
    L_I \le \frac{1024 \rho^2 \width \height \depth}{\ell^4} \left( 1+ \frac{7}{3} \pi \frac{\rho}{\width}
    \right)+ \rho \cdot o\left( \frac{\rho^3}{\ell^3}\right).
  \end{equation*}
\end{lemma}\smallskip
\begin{proof}
  A path visiting each cylinder once can be constructed by a sequence of
  passes, during which all cylinders in a row are visited by making left-to-right and then 
  right-to-left passes. This is done for all the rows of cylinders. In each row, there are at most $\lceil \width/\ell\rceil
  +1$ cylinders encountered in one pass. Hence, the cost
  of each pass is at most:
  \begin{equation*}
  L^\mathrm{pass}_I \le \width + 2 \ell + \rho \cdot
  o\left(\frac{\ell^2}{\rho^2}\right).
  \end{equation*} 
  In order to visit all cylinders in a row, the vehicle needs to make two passes through that row and the paths for these two passes are connected by a u-turn path whose length is $\frac{7}{3}\pi \rho$ + $\frac{\ell}{2}$. Therefore the length of the path required to visit all cylinders in one row is:
  \begin{equation*}
    L^\mathrm{row}_I \le 2 \width + \frac{9}{2} \ell + \frac{7}{3}\pi \rho + \rho \cdot
  o\left(\frac{\ell^2}{\rho^2}\right).    
\end{equation*}
  During the transition from one row to another, the vehicle needs to make a U-turn maneuver, in which the direction of
  travel is reversed, and the path moves to the next row, at distance equal
  to the diameter of the cylinder. Since the length of the shortest path to
  reverse the heading of the vehicle with co-located initial and final
  points is $(7/3) \pi \rho$, the length of the U-turn satisfies
  \begin{equation}
\label{eq:u-turn}
    L^\mathrm{U-turn}_I \le \frac{7}{3}\pi \rho + \frac{1}{2}w(\ell) \le
    \frac{7}{3} \pi \rho + \frac{\ell^2}{16 \rho} + \rho \cdot o\left(
      \frac{\ell^3}{\rho^3}\right).    
  \end{equation}  
  The total number of rows, i.e., the total number of rows of cylinders with
  non-empty intersection with $\domain$, satisfies
  \begin{equation*}
    N^\mathrm{row}_I \le \left\lceil \frac{2 \height}{w(\ell)} \right\rceil + 1
    \le \frac{16\rho H}{\ell^2} + 
    o\left(\frac{\rho}{\ell}\right).
  \end{equation*}  
During the transition from one row to another, the vehicle needs to make a U-turn maneuver whose length satisfies the same bound as in Eq. \eqref{eq:u-turn}
  The total number of layers of cylinders satisfies
\begin{equation*}
      N^\mathrm{layer}_I \le \left\lceil \frac{4 \depth}{w(\ell)} \right\rceil + 1
    \le \frac{32\rho \depth}{\ell^2} + 
    o\left(\frac{\rho}{\ell}\right).
\end{equation*}  
  A simple upper bound on the cost of closing the tour is given by 
  \begin{equation*}
    L^\mathrm{close}_I \le (W + 2 \ell) + (H + 2 w(\ell)) + (\depth + w(\ell))+ 2\pi \rho = W +
    H + \depth + 2\pi\rho + 2 \ell + \rho \cdot o(\ell/\rho).
  \end{equation*}  
  In summary, the total length of the path followed during the first sub-phase
  is
  \begin{align*}
    L_I & \le N^\mathrm{layer}_I \left(N^\mathrm{row}_I \left(L^\mathrm{row}_I +
      L^\mathrm{U-turn}_I\right) + L^\mathrm{U-turn}_I\right) + L^\mathrm{close}
    \\
    &\le \frac{1024 \rho^2 \width \height \depth}{\ell^4} \left( 1+ \frac{7}{3} \pi \frac{\rho}{\width}
    \right)+ \rho \cdot o \left( \frac{\rho^3}{\ell^3}\right).
\end{align*}
\end{proof}

Based on this calculation, we can estimate the length of the paths in
subsequent sub-phases.
 
\begin{eqnarray*}
L_{II} \le \frac{1024 \rho^2 \width \height \depth}{\ell^4} \left( 1+ \frac{7}{3} \pi \frac{\rho}{\width}
    \right)+ \rho \cdot o \left( \frac{\rho^3}{\ell^3}\right),\\
L_{III} \le \frac{512 \rho^2 \width \height \depth}{\ell^4} \left( 1+ \frac{7}{3} \pi \frac{\rho}{\width}
    \right)+ \rho \cdot o \left( \frac{\rho^3}{\ell^3}\right),\\
L_{IV} \le \frac{512 \rho^2 \width \height \depth}{\ell^4} \left( 1+ \frac{7}{3} \pi \frac{\rho}{\width}
    \right)+ \rho \cdot o \left( \frac{\rho^3}{\ell^3}\right),\\
L_{V} \le \frac{256 \rho^2 \width \height \depth}{\ell^4} \left( 1+ \frac{7}{3} \pi \frac{\rho}{\width}
    \right)+ \rho \cdot o \left( \frac{\rho^3}{\ell^3}\right).
\end{eqnarray*}

The length of path to execute the first phase is then the some of the path lengths for these five sub-phases.

\begin{lemma}[Path length at the first phase]
  \label{lemma:first-phase-3D}
  Consider a covering of the space with cylinders $\cyl(\ell)$. For any $\rho
  > 0$ and for any set of target points, the length $L_1$ of a path visiting once and only once each cylinder with a non-empty intersection with a rectangular box $\domain$ of width $\width$, height $\height$ and depth $\depth$ satisfies
  \begin{equation*}
    L_1 \le \frac{3328 \rho^2 \width \height \depth}{\ell^4} \left( 1+ \frac{7}{3} \pi \frac{\rho}{\width}
    \right)+ \rho \cdot o\left( \frac{\rho^3}{\ell^3}\right).
  \end{equation*}
\end{lemma}\smallskip
  
Since we increase the length of cylinders by a factor of two while doing the phase transtion from one phase to the another, the length of path for the subsequent $i^{\text{th}}$ phase is given by:
\begin{equation*}
    L_i \le \frac{3328 \rho^2 \width \height \depth}{16^i\ell^4} \left( 1+ \frac{7}{3} \pi \frac{\rho}{\width}
    \right)+ \rho \cdot o\left( \frac{\rho^3}{\ell^3}\right).
\end{equation*}

Finally, we can summarize these intermediate bounds into the main result of
this section. We let $\LenRCFA{\pointset}{\rho}$ denote the length of the
path computed by the \RecCylFillingAlgo for a point set~$\pointset$.
\begin{theorem}[Path length for the \RecCylFillingAlgo]
  \label{theorem:total-path-upper-bound-3D}
  Let $P\in\setpointsets{n}$ be uniformly randomly generated in the
  rectangle of width $W$, height $H$ and depth $\depth$. For any $\rho>0$, with high
  probability
  \begin{equation*}
    \lim_{n\to+\infty} \frac{\SOTSP{\pointset \subset \domain \subset \real^3}}{n^{4/5}} \,\le\,
    \lim_{n\to+\infty} \frac{ \LenRCFA{\pointset}{\rho} }{n^{4/5}} 
    \, \le \, \frac{3328}{15}\left(\frac{\pi}{16}\right)^{4/5} (\rho^2 W H \depth)^{1/5}.    
  \end{equation*}
\end{theorem}\smallskip

\begin{proof}
Clearly, 
\begin{align*}
L_\mathrm{RCFA} & = \sum_{i=1}^{\lceil \frac{\log[2]{n}+7}{5} \rceil} \left(\frac{3328 \rho^2 \width \height \depth}{16^i\ell^4} \left( 1+ \frac{7}{3} \pi \frac{\rho}{\width}
    \right)+ \rho \cdot o\left( \frac{\rho^3}{\ell^3}\right)\right) \\
& \le \frac{53248 \rho^2 \width \height \depth}{15\ell^4} \left( 1+ \frac{7}{3} \pi \frac{\rho}{\width}
    \right)+ \rho \cdot o\left( \frac{\rho^3}{\ell^3}\right).
\end{align*}
Recalling that $\ell = 2\left(\frac{16 \rho^2 WH \depth}{\pi n}\right)^{1/5} + o(n^{-1/5})$ for large $n$,
  the above can be rewritten as
  \begin{equation*}
    L_\mathrm{RCFA} \le \frac{3328}{15}\left(\frac{\pi}{16}\right)^{4/5} (\rho^2 W H \depth)^{1/5} \left( 1+ \frac{7}{3} \pi \frac{\rho}{\width}
    \right) n^{4/5} + o(n^{4/5}).
 \end{equation*}
\end{proof}

\begin{theorem}\textit{(Upper bound on the total time in $\real^3$)}
  \label{theorem:total-time-upper-bound-3D}
  Let $P\in\setpointsets{n}$ be uniformly randomly generated in the
  rectangular box of width $W$, height $H$ and depth $\depth$. For any
  double integrator \eqref{eq:second-order-model}, with high probability
  \begin{equation*}
    \lim_{n\to+\infty} \frac{\mathcal{T}_\mathrm{RecCCA}}{n^{4/5}} 
    \, \le \, 61 \left(\frac{W H \depth}{\umax^2\vmax}\right)^{1/5} 
    \left(1+ \frac{7 \pi \vmax^2}{3 W \umax} \right). 
  \end{equation*}
\end{theorem}\smallskip

\begin{remark}
  Theorems~\ref{theorem:lower-bounds}
  and~\ref{theorem:total-time-upper-bound-3D} imply that, with high
  probability, the \RecCCA is a $50\left(1+\frac{7\pi\vmax^2}{3\umax W}
  \right)$-factor approximation (with respect to $n$) to the optimal stochastic DITSP
  in $\real^3$ and that $\E[\SOTSP{\pointset \subset \domain \subset \real^3}]$
  belongs to $\Theta(n^{4/5})$.
\end{remark}

\section{The DTRP for double integrator}
\label{sec:DTRP}
We now turn our attention to the Dynamic Traveling Repairperson Problem
(DTRP) that was introduced in~\cite{DJS-GJvR:91} and that we here tackle
for a double integrator.

\subsection{Model and problem statement}
In the DTRP the double integrator is required to visit a dynamically
growing set of targets, generated by some stochastic process.  We assume
that the double integrator has unlimited range and target-servicing
capacity and that it moves at a unit speed with minimum turning radius
$\rho>0$.

Information about the outstanding targets representing the demand at time
$t$ is described by a finite set $n(t)$ of positions $\DD(t)$. Targets are
generated, and inserted into $\DD$, according to a time-invariant
spatio-temporal Poisson process, with time intensity $\lambda > 0$, and
uniform spatial density inside the region $\domain$, which we continue to
assume to be a rectangle for two dimensions and a rectangular box for three
dimensions.
Servicing of a target and its removal from the set $\DD$, is achieved when
the double integrator moves to the target position.  A control policy $\Phi$
for the DTRP assigns a control input to the vehicle as a function of its
configuration and of the current outstanding targets.
The policy $\Phi$ is a stable policy for the DTRP if, under its action
\begin{equation*}
  n_{\Phi} = \lim_{t\to+\infty} \E[n(t)|\;\dot{p}=\Phi(p,\DD)] < +\infty, 
\end{equation*}
that is, if the double integrator is able to service targets at a rate that
is, on average, at least as fast as the rate at which new targets are
generated.

Let $T_j$ be the time elapsed from the time the $j^{\text{th}}$ target is
generated to the time it is serviced and let $T_{\Phi}:=\lim_{j\to+\infty}
\E[T_j]$ be the steady-state system time for the DTRP under the policy
$\Phi$. (Note that if the system is stable, then it is known~\cite{LK:75}
that $n_{\Phi}=\lambda T_{\Phi}$.)  Clearly, our objective is to design a
policy $\Phi$ with minimal system time $T_\Phi$.

\subsection{Lower and constructive upper bounds}
In what follows, we design control policies that provide constant-factor
approximation of the optimal achievable performance. Consistently with the
theme of the paper, we consider the case of \emph{heavy load}, i.e., the
problem as the time intensity $\lambda\to+\infty$. We first provide lower bounds for the system time, and then present
novel approximation algorithms providing upper bound on the performance.

\begin{theorem}[Lower bound on the DTRP system time]
  \label{theorem:lower-bound-system-time}
  For a double integrator \eqref{eq:second-order-model}, the system time $T_{\mathrm{DTRP},2}$ and
  $T_{\mathrm{DTRP},3}$ for the DTRP in two and three dimensions satisfy
  \begin{align*}
    \lim_{\lambda\to\infty}\!\! \frac{T_{\mathrm{DTRP},2}}{\lambda^2}&
    \ge\frac{81}{32} \frac{W\!H}{\vmax \umax},
    \enspace
    \lim_{\lambda\to\infty}\!\! \frac{T_{\mathrm{DTRP},3}}{\lambda^4}
    \ge\frac{7813}{972} \frac{W\!H\!\depth}{\vmax \umax^2}.
  \end{align*}
\end{theorem}\smallskip

\begin{proof}
We prove the lower bound on $T_{\mathrm{DTRP},2}$; the bound on $T_{\mathrm{DTRP},3}$ follows on similar lines.
Let us assume that a stabilizing policy is available. In such a case, the number of outstanding targets approaches a finite steady-state value, $n^*$, related to the system time by Little's formula, i.e., $n^*=\lambda T_{\mathrm{DTRP},2}$. In order for the policy to be stabilizing, the time needed, on average, to service $m$ targets must be no greater than the average time interval in which $m$ new targets are generated. The average tim needed by the double integrator to service one target is no gretaer than the expected minimum time from an arbitrarily placed vehicle to the closest target; in other words, we can write the stability condition $\E[t^*(n^*)]\leq 1/\lambda$. A bound on the expected value of $t^*$ has been computed in the proof of Theorem~\ref{theorem:lower-bounds}, yielding
\begin{equation*}
\frac{3}{4}\left(\frac{6 W H}{\vmax \umax n}\right)^{1/3} \leq \E[t^*(n^*)]\leq 1/\lambda.
\end{equation*}
Using Little's formula $n^*=\lambda T_{\mathrm{DTRP},2}$, and rearranging, we get the desired result.
\end{proof}

We now propose simple strategies, the \BTA (for $\real^2$) and the \CFA (for $\real^3$), based on the concepts
introduced in the previous section.  The \BTA (\shortBTA) strategy consists of the following
steps:
\begin{enumerate}
\item Tile the plane with beads of length $\ell := \min\{
  C_\mathrm{BTA}/\lambda,4\rho\}$, where
  \begin{equation} 
    \label{eq:C_BTA}
    C_\mathrm{BTA}  =
    0.5241 \vmax  \left(1 + \frac{7\pi\rho}{3W}\right)^{-1}.
  \end{equation}
\item \label{step2} Traverse all non-empty beads once, visiting one target
  per non-empty bead. Repeat this step.
\end{enumerate}

The \CFA (\shortCCA) strategy is akin to the \shortBTA, where the region is covered with
cylinders constructed from beads of length $\ell := \min\{
C_\mathrm{CFA}/\lambda,4\rho\}$, where
\begin{equation*}
  C_\mathrm{CCA}  =
  0.1615 \vmax  \left(1 +\frac{7\pi\rho}{3W}\right)^{-1}.
\end{equation*} 
The policy is then to traverse all non-empty cylinders once, visiting one
target per non-empty cylinder.  The following result characterizes the
system time for the closed loop system induced by these algorithms and is
based on the bounds derived to arrive at
Theorems~\ref{theorem:total-time-upper-bound-2D} and
\ref{theorem:total-time-upper-bound-3D}.

\begin{theorem}[Upper bound on the DTRP system time]
  \label{theorem:DTRP}
  For a double integrator \eqref{eq:second-order-model} and $\lambda>0$, the \shortBTA and the \shortCCA are stable policies for the
  DTRP and the resulting system times $T_\mathrm{BTA}$ and $T_\mathrm{CFA}$ satisfy:
  \begin{gather*}
    \lim_{\lambda\to\infty}\! \frac{T_{\mathrm{DTRP},2}}{\lambda^2} \le
    \lim_{\lambda \to\infty}\! \frac{T_\mathrm{BTA}}{\lambda^2} \le 70.5 \frac{WH}{\vmax \umax} \left(1+\frac{7\pi\vmax^2}{3W\umax} \right)^3,
    \\
    \lim_{\lambda\to\infty}\! \frac{T_{\mathrm{DTRP},3}}{\lambda^4} \le
    \lim_{\lambda \to\infty}\! \frac{T_\mathrm{CFA}}{\lambda^4} \le 2\cdot 10^7
    \frac{W\!H\!\depth}{\vmax \umax^2}\!\left(\!1\! + \frac{7\pi\rho}{3W}\right)^5\!\!\!.
  \end{gather*}
  \end{theorem}\smallskip

\begin{proof}
We prove the upper bound on $T_{\mathrm{DTRP},2}$; the upper bound on $T_{\mathrm{DTRP},3}$ follows on similar lines.
  Consider a generic bead $B$, with non-empty intersection with $\domain$.
  Target points within $B$ will be generated according to a Poisson process
  with rate $\lambda_B$ satisfying
  \begin{equation*}
    \lambda_B = \lambda \frac {\Area(B \cap \domain)}{WH}\le \lambda\frac{
      \Area(B)}{WH} = \frac{C_\mathrm{BTA}^3}{16 \rho WH\lambda^2} +
    o\left(\frac{1}{\lambda^2}\right).    
  \end{equation*}
  The vehicle will visit $B$ at least once every $T_{\RecBTA,1}$ time units, where
  $T_{\RecBTA,1}$ is the bound on the time required to traverse a path of length $L_1$, as computed
  in Lemma~\ref{lemma:first_phase}. As a consequence, targets in $B$ will
  be visited at a rate no smaller than 
  \begin{equation*}
    \mu_B = \frac{C_\mathrm{BTA}^2 \vmax}{16 \rho WH \lambda^2}
    \left(1+\frac{7}{3}\pi \frac{\rho}{W}\right)^{-1} + o
    \left(\frac{1}{\lambda^2}\right).  
  \end{equation*}
  In summary, the expected time $T_\mathrm{B}$ between the appearance of a
  target in $B$ and its servicing by the vehicle is no more than the system
  time in a queue with Poisson arrivals at rate $\lambda_B$, and
  deterministic service rate $\mu_B$. Such a queue is called a $M/D/1$
  queue in the literature~\cite{LK:75}, and its system time is known to be
  \begin{equation*}
    T_{M/D/1} = \frac{1}{\mu_B} \left(1+\frac{1}{2}
      \frac{\lambda_B}{\mu_B-\lambda_B}\right).  
  \end{equation*}
  Using the computed bounds on $\lambda_B$ and $\mu_B$, and taking the
  limit as $\lambda\to+\infty$, we obtain
  \begin{equation}
    \label{eq:TB}
    \lim_{\lambda\to+\infty} \frac{T_\mathrm{B}}{\lambda^2} \le
    \lim_{\lambda\to+\infty} \frac{T_{M/D/1}}{\lambda^2} \le  
    \frac{16 \rho W H}{C^2_\mathrm{BTA}\vmax \left(1+\frac{7}{3}\pi
        \frac{\rho}{W}\right)^{-1}} \left(1 + \frac{1}{2}
      \frac{C_\mathrm{BTA}}{\vmax \left(1+\frac{7}{3}\pi \frac{\rho}{W}\right)^{-1}-
        C_\mathrm{BTA}}   \right). 
  \end{equation}
  Since equation~\eqref{eq:TB} holds for {\em any} bead intersecting
  $\domain$, the bound derived for $T_B$ holds for all targets and is
  therefore a bound on $T_\mathrm{DTRP,2}$. The expression on the right hand
  side of~\eqref{eq:TB} is a constant that depends on problem parameters
  $\rho$, $W$, and $H$, and on the design parameter $C_\mathrm{BTA}$, as
  defined in equation~\eqref{eq:C_BTA}. Stability of the queue is
  established by noting that $C_\mathrm{BTA} < \vmax (1+7/3\; \pi\;
  \rho/W)^{-1}$.  Additionally, the choice of $C_\mathrm{BTA}$ in
  equation~\eqref{eq:C_BTA} minimizes the right hand side of \eqref{eq:TB}
  yielding the numerical bound in the statement. We then substitute $\rho=\vmax^2/\umax$ to yield the final result.
\end{proof}

\begin{remark}
Note that the achievable performances of the \shortBTA and the \shortCCA provide a
constant-factor approximation to the lower bounds established
in Theorem~\ref{theorem:lower-bound-system-time}.
\end{remark}
\section{Extension to the TSPs for the Dubins vehicle}
In our earlier works \cite{KS-EF-FB:04l,KS-FB-EF:05j,KS-EF-FB:06h-tmp}, we
have studied the TSP for the Dubins vehicle in the planar case. In
\cite{KS-EF-FB:04l}, we proved that in the worst case, the time taken to
complete a TSP tour by the Dubins vehicle will belong to $\Theta(n)$. One
could shown that this result holds true even in $\real^3$. In
\cite{KS-EF-FB:06h-tmp}, the first known algorithm with strictly sublinear
asymptotic minimum time for tour traversal was proposed for the stochastic
DTSP in $\real^2$. This algorithm was modified in \cite{KS-EF-FB:06h-tmp}
to give a constant factor approximation to the optimal with high
probability.  This naturally lead to a stable policy for the DTRP problem
for the Dubins vehicle in $\real^2$ which also performed within a constant
factor of the optimal with high probability. The $\RecCCA$ developed in
this paper can naturally be extended to apply to the stochastic DTSP in
$\real^3$. It follows directly from Lemma~\ref{lem:feasible-curves} that in
order to use the $\RecCCA$ for a Dubins vehicle with minimum turning radius
$\rho$, one has to simply compute feasible curves for double integrator
moving with a constant speed $\sqrt{\rho \umax}$. Hence the results stated
in Theorem~\ref{theorem:total-time-upper-bound-3D} and
Theorem~\ref{theorem:DTRP} also hold true for the Dubins vehicle.

This equivalence between trajectories makes the \RecCCA the first known
strategy with a strictly sublinear asymptotic minimum time for tour
traversal for stochastic DTSP in $\real^3$. The fact that it performs
within a constant factor of the optimal with high probability and that it
gives rise to a constant factor approximation and stabilizing policy for
DTRP for Dubins vehicle in $\real^3$ is also novel.

\section{Conclusions}
\label{sec:conclusion}
In this paper we have proposed novel algorithms for various TSP problems
for vehicles with double integrator dynamics.  Future directions of
research include extensive simulations to support the results obtained in
this paper, study of centralized and decentralized versions of the DTRP,
and more general task assignment and surveillance problems for vehicles
with nonlinear dynamics.


{\small
  \bibliographystyle{ieeetr}%
  \bibliography{alias,Main,FB,New}
}

\end{document}